\documentclass[letterpaper, 10 pt, conference]{ieeeconf}  

\IEEEoverridecommandlockouts                              

\usepackage{graphicx} 
\usepackage{times} 

\usepackage{cite}       
\usepackage{adjustbox}  
\usepackage{multirow}   
\usepackage{makecell}  
\usepackage{mathtools, nccmath} 
\title{\LARGE \bf
Detecting agreement in multi-party dialogue: evaluating speaker diarisation versus a procedural baseline to enhance user engagement.
}

\author{Angus Addlesee$^{1}$, Daniel Denley$^{1}$, Andy Edmondson$^{1}$,  Nancie Gunson$^{1}$, Daniel Hernandez Garcia$^{1}$, \\ Alexandre Kha$^{1}$,  Oliver Lemon$^{1}$, James Ndubuisi$^{1}$,  Neil O’Reilly$^{1}$,  Lia Perochaud$^{1}$, \\ Raphaël Valeri$^{1}$ and Miebaka Worika$^{1}$%
\thanks{$^{1}$School of Mathematical and Computer Sciences,
        Heriot-Watt University, Edinburgh, Scotland
        }%
}

\begin{document}

\maketitle
\thispagestyle{empty}
\pagestyle{empty}

\begin{abstract}

Conversational agents participating in multi-party interactions face significant challenges in dialogue state tracking, since the identity of the speaker adds significant contextual meaning. It is common to utilise diarisation models to identify the speaker. However, it is not clear if these are accurate enough to correctly identify specific conversational events such as agreement or disagreement during a real-time interaction. This study uses a cooperative quiz, where the conversational agent acts as quiz-show host, to determine whether diarisation or a frequency-and-proximity-based method is more accurate at determining agreement, and whether this translates to feelings of engagement from the players.
Experimental results show that our procedural system was more engaging to players, and was more accurate at detecting agreement, reaching an average accuracy of 0.44 compared to 0.28 for the diarised system.

\end{abstract}

\section{INTRODUCTION}
\label{sec:intro}

The SPRING project aims to  develop social robots capable of multi-person interaction and communication in unstructured and populated environments \cite{gunson-etal-2022-visually}. One target environment is a hospital memory-clinic waiting room, which can be isolating despite the presence of others. Social robots are more effective than screen-based applications at promoting and aiding engagement \cite{sieinskaicebreaker}, so the project team designed a cooperative quiz intended for deployment on the ARI social robot \cite{cooper2020ari}. The quiz serves as an ice-breaking activity during which patients work together to identify answers, fostering constructive interaction, and briefly distracting patients from the stresses of their hospital visit. 

A key challenge for multi-party cooperative discussion is identifying when participants agree. Humans are very good at understanding who has spoken when, and at deriving contextual meaning from a conversation, but this remains challenging for artificial agents. To identify who has spoken, it is therefore common to use a form of diarisation \cite{RyantNeville2019TSDD}. However, accurate diarisation is difficult, depending on factors such as the quality of audio, the difference in voices and the ability of the model to accurately classify them \cite{1660031,addlesee-etal-2020-comprehensive, addlesee2023building}. It is therefore not clear if current diarisation models are more effective than other methods when developing a natural interaction with multiple parties. 

This paper proposes a conversational agent composed of a non-diarised, procedural system for detecting agreement between participants which is simple to implement, and consistent in operation. This is evaluated against a second system using the Google Cloud Speech to Text (STT) diarisation service. Our proposed system was more accurate at detecting agreement and provided more engaging interactions. Therefore, coupled with a simple quiz game, our method provides a benchmark against which to judge future methods. 

The following sections detail the system used for this baseline, the implementation of the cooperative quiz itself, and the results of a user participation study, comparing it to a modern diarised system.

 



\section{BACKGROUND RESEARCH}
\label{sec:Background Research}

Social robots are being used in an increasing number of settings including hospitals, like Baxter \cite{FitterNaomiT.2020EwBP} and the SPRING project \cite{Spring}, retail centres like MuMMER \cite{foster2019mummer}, pubs like JAMES \cite{James}, and museums like TINKER \cite{BickmoreTimothyW.2013Tara}.
We can observe robots such as JAMES \cite{James}, which can interact with several people to take orders, or the AMIGOS project \cite{Amigos} which was created to study the adaptation of interactions and emotions in the context of a group conversation between people and a social robot.
However, mastering the art of multi-person conversation is currently a very challenging capability for social robots to achieve \cite{addlesee2023data}.


It is important to consider a number of factors while developing a system that can interact with several users, including the ability to identify who is speaking, who is being addressed, when to respond, and what to respond with to a certain individual or group of people \cite{traum2004issues, ijcai2022p768,addlesee2023data}.


The goal of speaker diarisation is to answer the question ``WHO spoke when'' in an environment with several speakers and, commonly, various other noises \cite{RyantNeville2019TSDD, Dimitriadis2017DevelopingOS}. Speaker diarisation is the task of establishing how many speakers are present in a conversation and correctly identifying all segments for each speaker \cite{RyantNeville2019TSDD}.
Speech detection, segmentation, and clustering \cite{KOTTI20081091} serves as the foundation of diarisation, where the aim is to attribute a specific speaker for each part of the conversation \cite{1677976}. 
Recent work improves diarisation by using neural networks \cite{7953094, 8462628} or using Unbounded Interleaved-State Recurrent Neural Networks (UIS-RNN) \cite{zhang2019fully} but, despite these improvements, there still remain certain problems like speech overlap; when two speakers speak simultaneously, complicating diarisation \cite{1660031, addlesee2020comprehensive, liesenfeld2023timing}.

\section[Implementation]{METHODS}

\subsection{Design}
\label{sec:Design}
\begin{figure}[thpb]
  \centering
  \includegraphics[width=0.75\columnwidth,keepaspectratio]{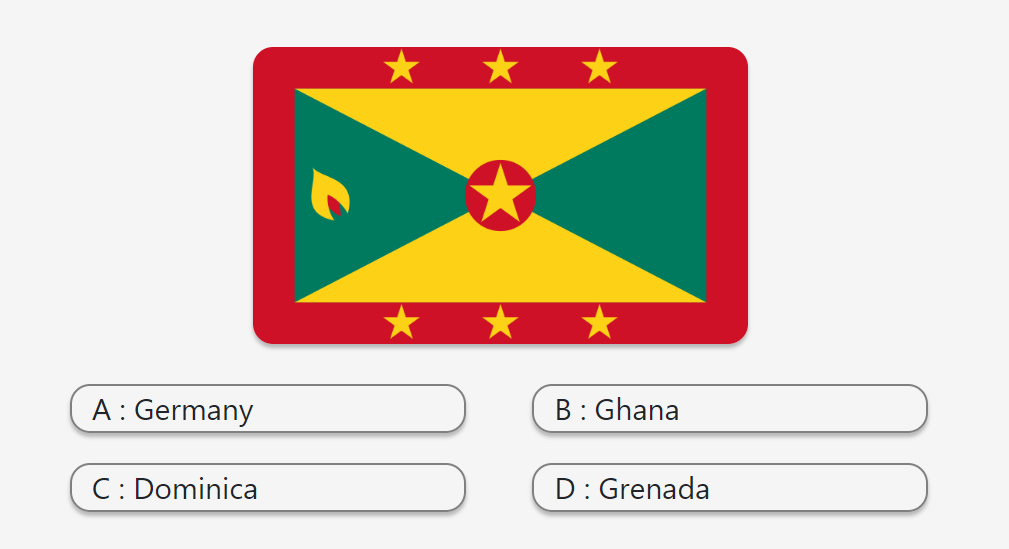}
  \caption[Flag Example]{The agent shows a flag and four options. 
  During this research, a laptop screen was used. The code enables integration with a social robot.}
  \label{fig:flag1}
\end{figure}

In order to provide light entertainment to patients in a hospital waiting room, the implemented cooperative quiz\footnote[2]{The code is available at {https://github.com/ddenley/Multi-Person-Quiz/}} is a flag identification game in which a pair of participants must identify what country the displayed flag belongs to, given four options. The flag images are selected at random from the ISO3166 country list \cite{ISO3166}, as are the four country names. A screen displays the flag (see Fig \ref{fig:flag1}), and the conversational agent asks the question, including the potential answers, using Text To Speech (TTS). The agent then listens to the participants' discussion to identify when countries are suggested, and when they have both agreed upon their final answer.


For this study, a laptop was used in place of an ARI robot to avoid confounding factors related to engagement with an embodied robot, such as the complications of implementing gestures and attention which obey human norms \cite{kahn_design_2008}.

To make the quiz interactions more natural, the agent can take conversational initiative under certain conditions. For example, if the participants disagree for several turns, the agent will offer them a clue. 
The system can suggest a clue, inquire whether the answer provided is the final answer, and, upon request, repeat the question, skip the question, or provide a clue.

\subsection{Architecture}
\label{sec:Architecture}
\begin{figure}[ht]
 \centering{\includegraphics[width=0.6\columnwidth,keepaspectratio]{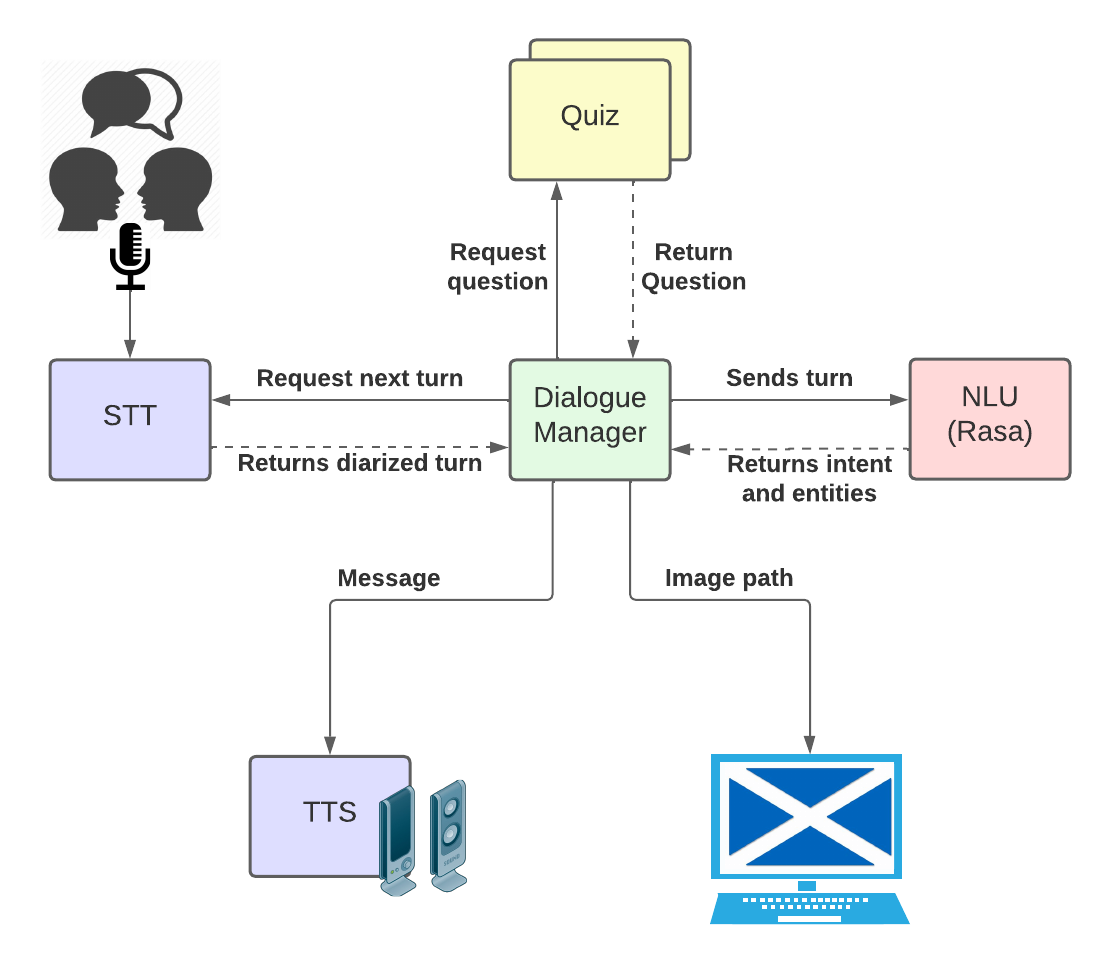}}
 \caption[System Architecture]{The system has a modular architecture, controlled centrally by its rule-based Dialogue Manager.}
 \label{fig:catFor}
\end{figure}

The system architecture is modular, controlled centrally by a rule-based Dialogue Manager (DM) as shown in Fig. \ref{fig:catFor}. Google Cloud is used for both the TTS and STT modules. Natural Language Understanding (NLU) is performed by Rasa Open Source \cite{bocklisch2017rasa}. 

Natural Language Generation (NLG) is a challenge, especially in a short game intended to be played several times in a row. In particular, repetitions significantly reduce engagement when compared with speech that has a very low rate of repetition \cite{see-etal-2019-makes}. To address this issue, the OpenAI GPT API was used to generate diverse versions of game host text and clues for all the countries (fact-checked for accuracy). The NLG module then sampled these utterances based on the decisions made by the Dialogue Manager.

\subsection{Natural Language Understanding (NLU)}
\label{sec:NLU}


NLU is performed on RASA using three main intents of interest, which are ``Give answer",``Agree",  and ``Disagree". Only one entity is considered, ``Country". The ``Give answer" intent encompasses any utterance containing an entity, whereas the other two describe simple ones, like ``Yes", ``No", ``I disagree", etc...

\subsection{Dialogue Manager (DM)}
\label{sec:DM}




The DM module controls the flow of information and the dialogue actions of the agent. Two versions of this module were created: the proposed baseline system without diarisation; and a diarised version for comparison, using the diarisation features of the Google Cloud STT service.
For both versions, when agreement is detected, a confirmation action is triggered asking players if the answer for which the agent has detected agreement is their final answer.



\subsubsection{Baseline: Decision-making without diarisation}
\label{sec:DM_no_diarisation}
The baseline approach is to count the number of \textit{give\_answer} intents for the same question, and check whether the last two answers are the same. The underlying intuition is that repetitions are likely to indicate agreement. If the number of \textit{give\_answer} intents is greater than a given threshold $N_A$, and if two following answers are the same, the system automatically detects  agreement as follows:
\begin{equation}
\label{eq:agree}
    \begin{aligned}
    \text{agreement = Y } \text{iff.} & (N_{answers} \geq N_A) \quad \text{and} \\ 
    &(answer_{t-1} = answer_t) \\
    \end{aligned}
\end{equation}
Additionally, if a player explicitly agrees with the answer of the other player - i.e. the intent is \textit{agree} and no questions other than the flag question have been asked by the system - this explicit agreement triggers the same action as detected implicit agreement. 

The value of $N_A$ was set through experimentation. The functionality is constrained by not having diarisation or negation, this being difficult for NLU modules to reliably manage \cite{hossain-etal-2022-analysis_negation_NLU}. With $N_A=2$, this leads to agreement false positives when one of the participants repeats the answer for any reason (e.g. to emphasise their certainty). The value of $N_A$ does need to be minimised for natural conversation, however, since it would be frustrating to require six (for example) repetitions from participants to detect their agreement. We found that a threshold of 3 provides suitable performance.


To illustrate the practicality of this threshold, an example dialogue is presented in Table \ref{tab:ex_dialogue}, The verification of agreement conditions, when $N_A=3$, can be found in Table \ref{tab:dialogue_conditions}.

\begin{table}[h!]
    \centering
    \caption{A real user dialogue with our system.}
     \label{tab:ex_dialogue}
    \begin{tabular}{ |l|l|l| }
        \hline
        \textbf{ID} & \textbf{Speaker}  & \textbf{Speech}\\
        \hline
        \hline
        1 & System   &  \makecell[l]{What flag is next in line to be shown? \\ Is this the flag of Christmas Island, Montserrat, \\ Czechia or Antigua and Barbuda? Now is the time \\to work  together and make your best guess. }\\
        \hline
        2 & P1   &  \makecell[l]{I'm pretty sure it is not Antigua and Barbuda. } \\
        \hline
        3 & P2   &  \makecell[l]{Yeah no way it's Antigua and Barbuda. } \\

        \hline
        4 & P1 & \makecell[l]{I would rather go for Christmas Island, what do \\ you think?} \\

        \hline
        5 & P2 & \makecell[l]{Sure, let's go for Christmas Island} \\

        \hline
        6 & System & \makecell[l]{So, is Christmas Island your final answer? }  \\
        
        \hline
    \end{tabular}
\end{table}
\begin{table}[h!]
    \centering
    \caption{Verification of the conditions needed to detect an agreement during the example dialogue ($N_A=3$)}
     \label{tab:dialogue_conditions}
    \begin{tabular}{ |l|c| c |c | }
        \hline
        \textbf{ID} & \textbf{N\_answers} & \textbf{Previous answer=Current answer} & \textbf{Detection?}\\
        \hline
        \hline
        1   &  0 & False & No\\
        \hline
        2   & 1 & False & No\\
        \hline
        3  & 2 & \textbf{True} & No\\

        \hline
        4 & \textbf{3} & False & No \\

        \hline
        5 & \textbf{4} & \textbf{True} & \textbf{Yes} \\
        
        \hline
    \end{tabular}
\end{table}

We can see in Table \ref{tab:dialogue_conditions} that using a threshold of 2 would fail in this situation, as the participants processed by elimination. Such a threshold would tend to over-detect agreements too early (here at the turn ID 3) and then would interrupt the discussion between the players. On the other hand, the threshold of 3 handled this situation (here detection at turn ID 5) and limited the number of additional turns needed by the system to detect agreement. 

It is important to note that the threshold only impacts one of the two agreement detection conditions in Equation \ref{eq:agree}. After three turns, our method only checks whether two consecutive answer entities match. We found that if pairs engaged in any process of elimination, it was at the beginning of their discussion. This finding, combined with the fact that there were only four given answer options, helped our threshold of 3 to further reduce false agreement detection. For example, see turn IDs 2 and 3 in Table \ref{tab:ex_dialogue}, our threshold prevented the elimination agreement from being detected as answer agreement. This reasoning is well adapted for quiz games with two participants, but might differ with additional players as the number of turns using negation statements could be higher. This would require further investigation to determine an optimal adaptive threshold depending on the use case.



\subsubsection{Comparison: Decision-making with diarisation}
\label{sec:DM_diarisation}
The detection of agreement when using the STT's diarisation feature is more straightforward. The label of each turn, transcribed to text by the STT and linked with its intent and a possible entity value returned by the NLU, allows the DM to store the current answer of each player. If the two current answers are the same, then the DM detects an agreement and triggers the confirmation action. This means that with accurate diarisation, each player could repeat their guess multiple times without triggering a false positive agreement detection. For instance, in a variation of Table \ref{tab:ex_dialogue} above, P1 could repeat ``Yes, Christmas Island'' to clarify their opinion to P2 without triggering a false positive agreement detection.

\subsection{Evaluation}
To compare the performance of the two systems, we evaluated each of them on two aspects: performance and player engagement. To determine whether a conversation is engaging, we retrieved metrics about turn-taking in the conversation. To get the users' opinions and feelings, we used a semi-structured interview \cite{LAZAR2017187}. In order to expose the performance of each system, we retrieved metrics to compare good agreement recognition, good disagreement recognition, and good intent understanding. To support this, we conducted an observation to record user reactions and behaviours, as well as potential unexpected behaviour of the system. Eight participants, 6 male and 2 female undergraduate students, were used for this evaluation. The participants were randomly divided into pairs and three games were played on each system by each pair. A game consisted of identifying three flags. To minimise bias, we alternated the sequence in which the DM systems were utilised for each test. 

\subsection{Experiment Protocol}
A laptop was placed on a table, with participants sitting side by side, facing the computer. They were verbally told the aim of the assessment and given instructions for the game and its features as detailed in the appendix \ref{sec:appendix_instruction}. The participants were not aware of the distinction between the two systems, simply referred to as system number 1 and system number 2.
Three games on system 1 were run and then three games on system 2, during the games we observed the users and system. The pair was then interviewed together after three games on each system. At the end of the evaluation, each system recorded the results of each game in a log file.

\section{RESULTS}

\subsubsection{Metrics}
\label{sec:metrics_eval}

For extrinsic evaluation, we compute the accuracy of agreement and disagreement detection of both systems. Explicit (NLU) and implicit (DM) agreements and disagreements are considered in these rates. They are defined as the number of correctly detected intents over their actual count. We consider that an agreement or disagreement is detected when the system asks for answer confirmation, or increments its disagreement counter ("disagree" intent or consecutive different answers detected); while actual (explicit and implicit) agreements and disagreements were annotated by the evaluators. Explicit Intent and Entity Recognition rates are also computed for more insight.

The non-diarised system considerably outperformed the diarised system on agreement detection, having an average success rate of \textbf{0.44}, compared to \textbf{0.28} for the diarised system. Full results gathered over the 24 games played are summarized in Tables \ref{tab:table_evaluation_ND} and \ref{tab:table_evaluation_D} in the Appendix, and illustrated in figure \ref{fig:success_rate}. 
In both systems, explicit intents and entity recognition performed equally, which was expected as they share the same NLU model. The NLU's intent recognition performed consistently, whereas the entity recognition was much more variable due to the STT (rates between $25\%$ and $100\%$). For example, ``Cyprus" was often confused with ``Cypress". Poor entity recognition makes the system unable to detect implicit agreement, explaining why some games lasted longer than others even though intents were understood. This effect was magnified in the diarised system, where speaker identification is additionally required for detecting agreement. Finally, we note that the disagreement recognition rate was often ill-defined and couldn't be computed because players discussed without necessarily disagreeing. 

\subsubsection{Observation}
\label{sec:obs}

In general, players were engaged by the quiz. They discussed their ideas, especially for the most challenging questions, used game features (e.g. asking for clues) and were motivated to find the correct answer. When the systems were not responsive, however, they became frustrated with the game as they did not know what they could say to get a response \cite{10.1145/3173574.3174214}. Most groups tried to repeat the same answer several times to get the system to take their answer and agreement into account, but most of the time they just tried to get any reaction from it. For example, one group just said \textit{``Hello!"}. Today's conversational agents usually interrupt users \cite{addlesee2023understandingsparql}, so it was interesting to find that our attempts to intelligently listen to a multi-party conversation raised the opposite issue.

The games played were not fluent and natural. The players had to repeat their answers several times, and sometimes had to explicitly say that they agree. The diarised system could not reliably distinguish participants' voices, resulting in failed agreement detection. The non-diarised system failed to detect agreement in the easiest questions as the correct answer was given in the first two turns, before the threshold.

The clue was very useful sometimes for the challenging questions and added a new point of discussion for the players. Moreover, the detection of the clue request by the system was reliable.

Finally, we identified some unexpected behaviours from the players, not covered by the system. For example, two of the four groups tried to use the names of the options given to the players - like answer \textit{A} - before understanding that the system does not handle this feature. We also noticed that some participants interrupted the robot when asking the third part of the flag question. The first part asks the players what the displayed flag is, the second part lists the different choices, and the third part is about the general principle of the quiz like ``Collaborate together to find the answer to this question''.

\subsubsection{Interview}
\label{sec:interview}

The first question asked the player if they found the quiz engaging and why. Most of the players concurred that the game is engaging mainly because of its original features.
However, one player pointed out that they did not find the quiz engaging because the system was often unresponsive. 

The second question focused more precisely on the feeling of the players regarding the performance of the system, asking if they felt their answers and agreements were understood. All the players noticed that the system did not always identify agreement, often requiring several repetitions of their answer or an explicit statement of agreement, for example by saying \textit{``Yes, I agree"}. One player also mentioned that it would be useful to receive more interactive feedback about what the system has understood. The user could then adapt their behaviour accordingly, by adapting their pronunciation, for example.

Finally, the last question asked the players which system - 1 or 2 - they preferred. All groups of players were adamant in their answers, with three of the four groups preferring the procedural method, with group 3 preferring the diarised version. Interestingly, this one pair was the only group made up of a mixed male/female pair, whose voices were likely the most dissimilar. They all said that the system they preferred was much more responsive than the other, in which they had to repeat their answer several times before getting a reaction from the system. 

\section{DISCUSSION}

As stated in the introduction, the key purpose of the game is to encourage interaction between people. A general concern related to procedural systems is that they will feel unnatural and stilted, affecting engagement. However, as shown in Sections \ref{sec:obs} and \ref{sec:interview}, the procedural baseline was found to be the more engaging and a viable method in itself. The key weakness identified in the presented system is in cases of immediate agreement between the two participants, where it requires a third statement of agreement to trigger detection.

While the diarised system was found to be less engaging, this study did not find that non-diarised systems are inherently superior. Rather, that the current diarisation techniques still struggle to differentiate speakers in noisy environments or where voices not sufficiently dissimilar \cite{addlesee2020comprehensive}. For example, the best results obtained by the diarised system were with a male and female who had very different voices and accents who, during interview, reported the interaction to be natural and engaging. Illustrating how unusual this sufficient dissimilarity is, only one participant pair had this experience. Our study was small, so it is not possible to draw conclusions, but it may suggest that future diarised systems could provide a natural user experience. 

Looking to future systems, by releasing this new game and baseline system, we aim to provide a benchmark that encourages future advancements in the field by enabling an easily implemented way to measure against a known viable system. Researchers and practitioners can build on this foundation to evaluate their own systems, adapt them for deployment on platforms such as the ARI robot, for which this project is planned, and contribute to ongoing advances in multi-party spoken dialogue systems.

\clearpage
\bibliographystyle{unsrt}
\bibliography{anthology, custom}

\clearpage          

\appendices
\section{Details of instructions given to participants}
\label{sec:appendix_instruction}

At the beginning of each test, we introduced our project with a short description: "Hello! Our project has the goal of helping people break the ice and keep them entertained while they wait in a waiting room. We've developed a game where the computer is the game master, and you are the players."

Next, we outlined the tasks for them:
"Your task is to test two different systems. In each system, you'll play three games. To win a game, you need to get three correct answers."

After that, we described the game itself:
"In the game, a flag will be shown, and the objective is to work together to pick the correct answer from four options provided. If you're stuck, you can ask for a hint or skip the question."

We intentionally kept our explanations concise, as we wanted to assess how user-friendly our system is without extensive instructions.

\section{Agreement recognition rate for the different groups used for evaluation}
\begin{figure}[thpb]
  \centering
  \includegraphics[width=1\columnwidth,keepaspectratio]{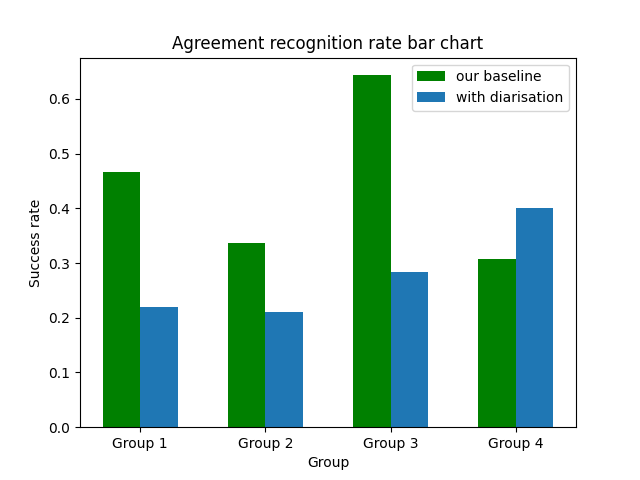}
  \caption[Agreement rate]{Bar chart of the agreement recognition rate for the different groups used for evaluation.}
  \label{fig:success_rate}
\end{figure}

\clearpage
\section{Extrinsic evaluation metrics}
\label{sec:appendix_extrinsic}

\begin{table}[h!]
\begin{center}
\caption{Evaluation for the procedural baseline system}
\label{tab:table_evaluation_ND}
\begin{adjustbox}{width=1\textwidth}
\begin{tabular}{ |c|c||c|c|c|c|c| } 
\hline
 \textbf{Group} & \textbf{Game} & \textbf{Nb turns} & \textbf{Agreement recognition rate} & \textbf{Disagreement recognition rate} & \textbf{Explicit intent recognition rate} & \textbf{Entity recognition rate} \\
\hline
\hline
\multirow{3}{4em}{\textbf{Group 1}} & G1 & 23 & 0.44 & N/A & 1 & 0.6  \\ 
& G2& 27 & 0.46 & N/A & 0.89 & 0.83  \\ 
& G3& 23 & 0.5 & N/A & 0.87 & 0.56  \\ 
\hline
\multirow{3}{4em}{\textbf{Group 2}} & G4 & 23 & 0.42 & N/A & 0.83 & 0.73  \\ 
& G5& 19 & 0.35 & N/A & 0.79 & 0.92  \\ 
& G6& 27 & 0.24 & N/A & 0.89 & 0.75  \\ 
\hline
\multirow{3}{4em}{\textbf{Group 3}} & G7 & 31 & 0.53 & 0.12 & 0.87 & 0.88  \\ 
& G8& 15 & 0.83 & 0.19 & 1 & 1  \\ 
& G9& 24 & 0.57 & 0.43 & 0.88 & 0.55  \\ 
\hline
\multirow{3}{4em}{\textbf{Group 4}} & G10 & 19 & 0.38 & N/A & 0.89 & 0.6  \\ 
& G11& 23 & 0.27 & N/A & 0.74 & 0.73  \\ 
& G12& 22 & 0.27 & N/A & 0.77 & 0.7  \\ 
\hline
\hline
\textbf{Mean} & & \textbf{23}& \textbf{0.44} & \textbf{N/A} & \textbf{0.87} & \textbf{0.74} \\
\hline
\end{tabular}
\end{adjustbox}
\end{center}
\end{table}


\begin{table}[h!]
\begin{center}
\caption{Evaluation for the diarised system}
\label{tab:table_evaluation_D}
\begin{adjustbox}{width=1\textwidth}
\begin{tabular}{ |c|c||c|c|c|c|c| } 
\hline
 \textbf{Group} & \textbf{Game} & \textbf{Nb turns} & \textbf{Agreement recognition rate} & \textbf{Disagreement recognition rate} & \textbf{Explicit intent recognition rate} & \textbf{Entity recognition rate} \\
\hline
\hline
\multirow{3}{4em}{\textbf{Group 1}} & G1 & 37 & 0.11 & 0.5 & 0.9 & 0.71  \\ 
& G2& 29 & 0.3 & N/A & 0.97 & 0.56  \\ 
& G3& 37 & 0.25 & N/A & 1 & 0.25  \\ 
\hline
\multirow{3}{4em}{\textbf{Group 2}} & G4 & 32 & 0.1 & 0.11 & 0.69 & 0.75  \\ 
& G5& 33 & 0.16 & N/A & 0.79 & 0.81 \\ 
& G6& 26 & 0.37 & N/A & 0.85 & 1  \\ 
\hline
\multirow{3}{4em}{\textbf{Group 3}} & G7 & 30 & 0.3 & N/A & 0.73 & 0.43  \\ 
& G8& 34 & 0.21 & N/A & 0.74 & 0.84  \\ 
& G9& 23 & 0.34 & N/A & 0.91 & 0.73  \\ 
\hline
\multirow{3}{4em}{\textbf{Group 4}} & G10 & 39 & 0.19 & 0.14 & 0.88 & 0.67  \\ 
& G11& 14 & 0.75 & N/A & 1 & 1 \\ 
& G12& 34 & 0.26 & 0.38 & 0.94 & 0.71  \\ 
\hline
\hline
\textbf{Mean} & & \textbf{30.67}& \textbf{0.28} & \textbf{N/A} & \textbf{0.79} & \textbf{0.71} \\
\hline
\end{tabular}
\end{adjustbox}
\end{center}
\end{table}

\clearpage

\clearpage

\end{document}